\documentclass[journal]{IEEEtran}
\usepackage{graphicx}
\usepackage{amsmath}
\usepackage{amssymb}
\usepackage{cite}
\usepackage{amsmath}
\usepackage{algorithmic}
\usepackage{array}

\ifCLASSOPTIONcompsoc
 \usepackage[caption=false,font=normalsize,labelfont=sf,textfont=sf]{subfig}
\else
 \usepackage[caption=false,font=footnotesize]{subfig}
\fi

\usepackage{url}

\begin{document}

\title{Explainable Deep Reinforcement Learning for UAV Autonomous Navigation}

\author{Lei He, ~\IEEEmembership{Student Member, ~IEEE}, Aouf Nabil, ~\IEEEmembership{Member, ~IEEE}, Bifeng Song  

\thanks{Lei He and Bifeng Song are with the School of Aeronautics, Northwestern Polytechnical University, Xi’an 710072, China (email: heleidsn@gmail.com)}
\thanks{Aouf Nabil is with Department of Electrical and Electronic Engineering, City University of London, EC1V 0HB, UK}}

\maketitle

\begin{abstract}
Autonomous navigation in unknown complex environment is still a hard problem, especially for small Unmanned Aerial Vehicles (UAVs) with limited computation resources. In this paper, a neural network-based reactive controller is proposed for a quadrotor to fly autonomously in unknown outdoor environment. The navigation controller makes use of only current sensor data to generate the control signal without any optimization or configuration space searching, which reduces both memory and computation requirement. The navigation problem is modelled as a Markov Decision Process (MDP) and solved using deep reinforcement learning (DRL) method. Specifically, to get better understanding of the trained network, some model explanation methods are proposed. Based on the feature attribution, each decision making result during flight is explained using both visual and texture explanation. Moreover, some global 
analysis are also provided for experts to evaluate and improve the trained neural network. 
The simulation results illustrated the proposed method can make useful and reasonable explanation for the trained model, which is beneficial for both non-expert users and controller designer. Finally, the real world tests shown the proposed controller can navigate the quadrotor to  goal position successfully and the reactive controller performs much faster than some conventional approach under the same computation resource. 


\end{abstract}


\IEEEpeerreviewmaketitle

\section{Introduction}

\IEEEPARstart{U}{nmaned} Aerial Vehicles (UAVs) have been widely used in many application, such as good delivery, emergency surveying and mapping. 
Autonomous navigation in the large unknown complex environment is an essential capability for these UAVs to operate more intelligent and safety. 

In general, there are two main solutions for UAV obstacle avoidance. 
The first solution relies on the state estimator using VIO or SLAM, then generate safety trajectories using optimization method \cite{liu2017planning, zhou2019robust} or searching-based method. 
It's a cascade process include mapping, localization planning and control. This kind of method can generate nearly optimal trajectories for some optimization objectives such as safety and smoothness, however, they require lots of computation and memory to store the map and run the optimization algorithms every step. 
In addition, these techniques also suffer from high drift and noise, impacting the quality of both localization and the map used for planning. 
Another solution is using a reactive control method, which can generate control command from the perception information directly \cite{paschall2017fast, escobar2018r}.
This method require less computation and memory resources because the control signal is obtained using only one forward calculation and it doesn't need to maintain the map during flight. This property makes it promising for some micro UAV with size, weight and power (SWaP) constraints. The weakness is that it is always non-optimal because of the lacking of global information. Also, the design of the reactive policy always relies on the expert experiment. 

Reactive UAV navigation based on only current information can be formulated as a sequential decision-making problem. Some researchers modelled this problem as a Markov decision process (MDP) and solved using reinforcement learning (RL) methods.
For example, Ross \emph{et~at} \cite{ross2013learning} build and Imitation learning (IL)-based controller using a small set of human demonstrations and achieved a good performance in natural forest environments. Imanberdiyev \emph{et~at} \cite{imanberdiyev2016autonomous} developed a high-level control method for autonomous navigation of UAVs using a novel model-based reinforcement learning method, TEXPLORE. He \emph{et~al} \cite{He2020Integrated} combine bio-inspired monocular vision perception method with a deep reinforcement learning (DRL) reactive local planner to address the UAV navigation problem. They also proposed learning from demonstration method to speed up the training process \cite{he2020deep}. Wang \emph{et~al} \cite{wang2019autonomous} formulated the navigation problem as a partially observable Markov decision process (POMDP) and solved by a novel online DRL algorithm. He also invested the sparse reward situation using a learn with help (LwH) method \cite{wang2020deep}.
Comparing to the traditional rule-based reactive controller, the control policy trained by DRL can get near optimal action in the training environment. Also, rely on the powerful feature extraction capacity of deep neural network (DNN), the trained policy can get feature autonomously without human design, always can get better performance. 

Although DRL method can get excellent performance, an enormous problem is that deep learning methods turn out to be uninterpretable “black boxes,” which create serious challenges to the Artificial Intelligence (AI) system based on neural network \cite{goebel2018explainable}. This problem falls with the so-called eXpalinable AI (XAI) filed. Arrieta \emph{et~al} gives a review of XAI, include concepts, taxonomies, opportunities and challenges toward responsible AI \cite{arrieta2020explainable}. 

Comparing to the burst of XAI research in supervised learning, explainability for RL is hardly explored \cite{heuillet2020explainability}. Juozapaitis \emph{et~al} \cite{juozapaitis2019explainable} explain the RL agent using reward decomposition. This approach decomposes reward into sums of semantically meaningful reward types so that actions can be compared in terms of trade-offs among the types. 
Reward deposition is also used in strategic tasks such as StarCraft II \cite{pocius2019strategic}. Jung Hoon Lee \cite{lee2019complementary} proposed a method to derive a secondary comprehensible agent from NN-based RL agent, the decision makings are based on simple rules. Beyret \emph{et~at} \cite{beyret2019dot} proposed a explainable RL for robotic manipulation. They presented a hierarchical DRL system include both low-level agent handling actions and high-level agent learning the dynamics and the environment. The high-level agent is used to interpret for the human operator. Madumal \emph{et~at} \cite{madumal2019explainable} use causal models to derive causal explanations of the behaviour of model-free reinforcement learning agent. A structural causal model is learned during the reinforcement learning phase. The explanations of behaviour are generated based on the counterfactual analysis of the causal model. They also introduced a distal explanation model that can analyse counterfactual and opportunity chains using decision trees and causal models \cite{madumal2020distal}. 

Explainability is critical and essential for DRL-based UAV navigation system. On the one hand, it's useful for non-expert users to know the reason why the controller turn right rather than turn left when it facing an obstacle. On the other hand, it also benefits the network and controller designer to know the decision making progress and do some adjustment to improve network performance.

In this paper, an end-to-end neural network is proposed to address the UAV reactive navigation problem in the complex unknown environment for small UAVs with SWaP constraints. The network is trained using DRL method in a high-fidelity simulation environment. Then, a post-hoc explanation method is proposed to provide explainable information of the trained network.
Comparing to the transparent model methods, post-hoc methods can provide explanations of an RL policy after its training, which keeps the model performance.  
To get better understanding of the trained network, both visual and textual explanations to each model output are provided as local explanation for non-expert users. Moreover, some global explanations are also provided for experts to analyze and improve the network. 

Our main contributions can be summarised as follows:
\begin{itemize}
    \item A DNN-based reactive controller for UAV navigation is learned using DRL method which can be used by some small UAVs with limited computation resources or some scenarios need very rapid reaction to the environment changes. 
    \item A novel CNN attention visualization method as well as texture explanation are provided based on a more fair feature attribution than gradients.
    \item Some local and global explanations are provided for non-expert users and expert to diagnose the trained DNN model. 
    \item Real world experiments are carried out to validate the trained network and show the computation efficiency of our reactive controller comparing a conventional searching-based approach.  
\end{itemize}

\section{Preliminaries}
\subsection{MDP and DRL}
In this work, the navigation and obstacle avoidance problem is formulated using MDP. An MDP is defined by a tuple $<S,A,R,P,\gamma>$, which is consists of a set of states $S$, a set of actions $A$, a reward function $R(s,a)$, a transition function $P(s'|s,a)$, and a discount factor $\gamma \in (0,1)$. In each state $s \in S$, the agent takes an action $a \in A$. After executing the action $a$ in the environment, the agent receives a reward $R(s,a)$ and reaches a new state $s'$, determined from the probability distribution $P(s'|s,a)$. 

Solutions for MDPs with finite state and action spaces can be found through a variety of methods such as dynamic programming, especially when the transition probabilities are given. However, in most of the MDPs, the transition probabilities or the reward functions are not available. In this situation, to solve the MDP, we need to interact with the environment to get some inner information, which is RL method. The goal of RL is to find a policy $\pi$ mapping states to actions that maximizes the expected discounted total reward over the agent's lifetime. This concept is formalized by the action value function: $Q^{\pi}(s,a)=\mathbb{E}^{\pi} \left[\sum^{T}_{t=0} \gamma^{t}R(s_t,a_t)\right]$, where $\mathbb{E}^{\pi}$ is the expectation over the distribution of the admissible trajectories $(s_0,a_0,s_1,a_1,\dots)$ obtained the policy $\pi$ starting from $s_0=s$ and $a_0=a$. The action value function can be defined by a tabular mapping of discrete inputs and outputs. However, this is tabular mapping is limiting for continuous states or an infinite/large number of states. Different from the traditional RL algorithms, DRL algorithms uses DNN to approximate the action value function, as opposed to tabular functions, makes it can deal with complex problem with infinite/large number of states. 

\subsection{Explainable AI and Explainable RL}
In recent years, AI has achieved a notable momentum and lies at the core of many activity sectors. However, because of the black-box property of the deep neural network model, the demand for transparency is increasing from the various of stakeholders in AI. In order to avoid limiting the effectiveness of the current generation of AI system, XAI techniques are proposed to produce more explainable models while maintain a high level of learning performance and enable humans to understand and trust the emerging generation of artificially intelligent partners \cite{arrieta2020explainable}. 

There are two kind of methods to increase the transparency of AI models, a widely accept classification is that using transparency models or using post-hoc XAI techniques. A model is considered to be transparent if by itself it is understandable, such as linear regression, decision trees, rule-based models, etc. This kind of model is usually simple enough to be understand by humans or designed by some manually set rules. However, as the increase of the model prediction accuracy, more and more models using deep and complex neural network appear. This kind of models cannot be easily understand by humans directly. Thus, post-hoc XAI techniques are important to handle such complex models to provide us some inner view. 

The success of DRL could augur an imminent arrival in the industrial world. However, like many Machine Learning algorithms, RL algorithms suffer from a lack of explainability. Although a large set of XAI literature is emerging to explain the DNN output, assessing how XAI techniques can help understand models beyond classification tasks, e.g. for reinforcement learning (RL), has not been extensively studied \cite{heuillet2020explainability}. Furthermore, DRL models are usually complex to debug for developers as they rely on many factors, such as environment, reward function, observation and even the algorithms used for training the policy. Thus, there is an urgent demand for explainable RL (XRL). 

\subsection{Feature Attribution}
Feature attribution is a common method to analysis trained DNN model.
Formally, suppose we have a function $F:R^n \to [0,1]$ that represents a deep neural network and an input $x=(x_1, \dots, x_n) \in R^n$. An attribution of the prediction at input $x$ relative to a baseline input $x'$ is a vector $A_F(x, x')=(a_1, \dots, a_n) \in R^n$ where $a_i$ is the contribution of $x_i$ to the prediction $F(x)$. 
There are two different types of feature attribution algorithms: Shapley-value-based algorithm and gradient-based algorithm. There is a fundamental difference between these two algorithm types.

Shapley value is a classic method to distribute the total gains of a collaborative game to a coalition of cooperating players. It is a fair way to attribute the total gain to the players based on their contributions.
For ML models, we formulate a game for the prediction at each instance. We consider the “total gains” to be the prediction value for that instance, and the “players” to be the model features of that instance. The collaborative game is all of the model features cooperating to form a prediction value. A Shapley-value-based explanation method tries to approximate Shapley values of a given prediction by examining the effect of removing a feature under all possible combinations of presence or absence of the other features. 
Shapley values are the only additive feature attribution method that satisfies the desirable properties of local accuracy, missingness, and consistency. However, exact Shapley value computation is exponential in the number of features. 

Besides the Shapley values, gradients can also used as the feature attribution. A gradient-based explanation method tries to explain a given prediction by using the gradient of the output with respect to the input features. However, the problem with gradients is that they break sensitivity, a property that all attribution methods should satisfy. For example, consider a one variable, one ReLU network, $f(x) = 1 - \text{ReLU} (1 - x)$. Suppose the baseline is $x=0$ and the input is $x=2$. The output changes from 0 to 1, but the gradient is zero at $x=2$ because $f$ becomes flat after $x=1$, so the gradient method gives attribution of 0 to $x$. This phenomenon has been reported in \cite{shrikumar2016not}. To address this problem, Sundararajan \emph{et~al} \cite{sundararajan2017axiomatic} proposed Integrated Gradients (IG) algorithm. However, this algorithm requires computing the gradients of the model output on a few different inputs (typically 50) between current feature value and baseline value.

\subsection{SHAP and DeepSHAP}
SHAP (SHapley Additive exPlanations), proposed by Lundberg and Lee \cite{lundberg2017unified}, can assigns each feature an importance value for a particular prediction. For a simple linear regression problem, the predictions can be written as:
\begin{equation}
    \hat{y_i} = b_0 + b_1 x_{1i} + \cdots + b_d x_{di}
\end{equation}
where $\hat{y_i}$ is the \textit{i}-th predicted response, ${x_{1i},\dots,x_{di}}$ are the features of current observation, and ${b_0, \dots, b_d}$ are the estimated regression coefficients. If the features are independent, the contribution of the \textit{k}-th feature to the predicted response $\hat{y_i}$ can be unambiguously expressed as $b_k x{ki}$ for $k=1, \dots, d$.

SHAP is a generalization of this concept to more complex neural network models. We define the following:
\begin{itemize}
    \item $F$ is the entire set of features, and $S$ denotes a subset.
    \item $S \cup i$ is the union of the subset $S$ and feature $i$. 
    \item $E[f(X)|X_s=x_s]$ is the conditional expectation of model $f(\dot)$ when a subset $S$ of features are fixed at the local point $x$.
\end{itemize}
Then, the SHAP value is defined to measure the contribution of the \textit{i}-th feature as
\begin{equation}
    \Phi_i = \sum_{S \subseteq F \setminus \left\{ i \right\}} \frac{|S|!(|F|-|S|-1)!}{|F|!} [f_{S \cup \left\{i\right\}}(x_{S \cup \left\{i\right\}}) - f_S(x_S)]
\end{equation}

SHAP values are proved to satisfy good properties such as fairness and consistency on attributing importance scores to each feature. But the calculation of SHAP values is computationally expensive. In our case, we use Deep SHAP, which is a model-specific method to improve computational performance through a connection between Shapley values and DeepLIFT \cite{shrikumar2017learning}. 

DeepSHAP \cite{chen2019explaining} is a framework for layer-wise propagation of Shapley values that builds upon DeepLIFT \cite{shrikumar2017learning}. If we define including an input as setting it to its actual value instead of its reference value, DeepLIFT can be thought of as a fast approximation method of the Shapley values. If our model is fully linear, we can get exact SHAP values by summing the attributions along all possible paths between input $x_i$ and the model's output $y$. However, in our network, for example fully connected network, there are non-linear activation function applied after the linear part, such as ReLU, tanh or sigmoid operations. To deal with the non-linear part, DeepLIFT provided the Rescale rule and the RevealCancel rule. Passing back nonlinear attributions linearly is an approximation, but there are two main benefits: 1) fast computation using only one backward pass and 2) a guarantee of local accuracy.

\section{DRL-based UAV navigation}
In this section, we introduce a DRL-based reactive controller to solve the UAV navigation problem in unknown environment. In contrast to conventional simultaneous localization and mapping-based method, we control the UAV only according to the current sensor data. This kind of controller can make quick reaction in the complex environment, which is beneficial to the small UAVs with limited computation resources. 

Reactive navigation in unknown environment can be treated as a sequential decision problem. In each time step, the current sensor information is used to generate the final control signal. This means the action $a$ depends only on the current state $s$ and the next state $s'$ depends on the current state $s$ and the decision maker's action $a$. This kind of problem can be modelled as a MDP after defining a corresponding reward function $R_a(s,s')$. 

\subsection{Problem Formulation}
\begin{figure*}[t]
	\centering
	\includegraphics[width=0.8\textwidth]{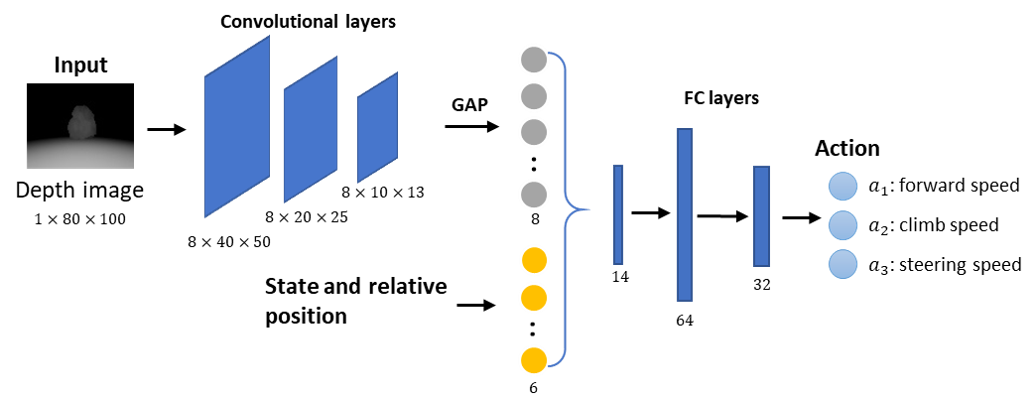}
	\caption{Network architecture of our control policy. The input is raw depth image and UAV states such as current speed and relative position to the goal. The features in the Depth image is extracted using CNN. Then global average pooling layer is used to get the intensity of each visual feature and then feed to the fully connected network combined with state features. The outputs are 3 control command includes forward, climb and steering speed.}
	\label{fig_network_arch}
\end{figure*}

Suppose the UAV takes off from a departure position in a 3-D environment, which is denoted as $(x_0, y_0, z_0)$ in the Earth-fixed coordinate frame, and targets at flying to a destination that is denoted as $(x_d, y_d, z_d)$.
The observation or the state at time $t$ consists of both raw depth image and UAV state features: $o_t = [o_\text{depth}^t, o_\text{state}^t]$. The state feature consists of the relative position to goal and current velocity information: $o_\text{state}^t = [d_{xy}^t, d_z^t, \xi^t, v_{xy}^t, v_z^t, \phi^t$], where $d_{xy}^t$ and $d_z^t$ denote the distance between the UAV's current position and the destination position in x-y plane and z axis, $\xi^t$ is the relative angle between UAV current first-perspective direction to the destination position, $v_{xy}^t$ and $v_z^t$ are the UAV current speed and $\phi^t$ is the steering angular speed. 
Action $a = [v_{xy}^\text{cmd}, v_z^\text{cmd}, \phi^\text{cmd}]$ generated from the policy network $\pi(s)$ consists of 2 linear velocity and 1 angular velocity. These actions are passed to the low-level controller as velocity setpoint command to achieve the navigation. The network architecture of the navigation network is shown in Fig. \ref{fig_network_arch}.

\subsection{Training Environment and Setting}
\begin{figure}
	\centering
	\subfloat[AirSim simulation environment]{\includegraphics[width=0.24\textwidth]{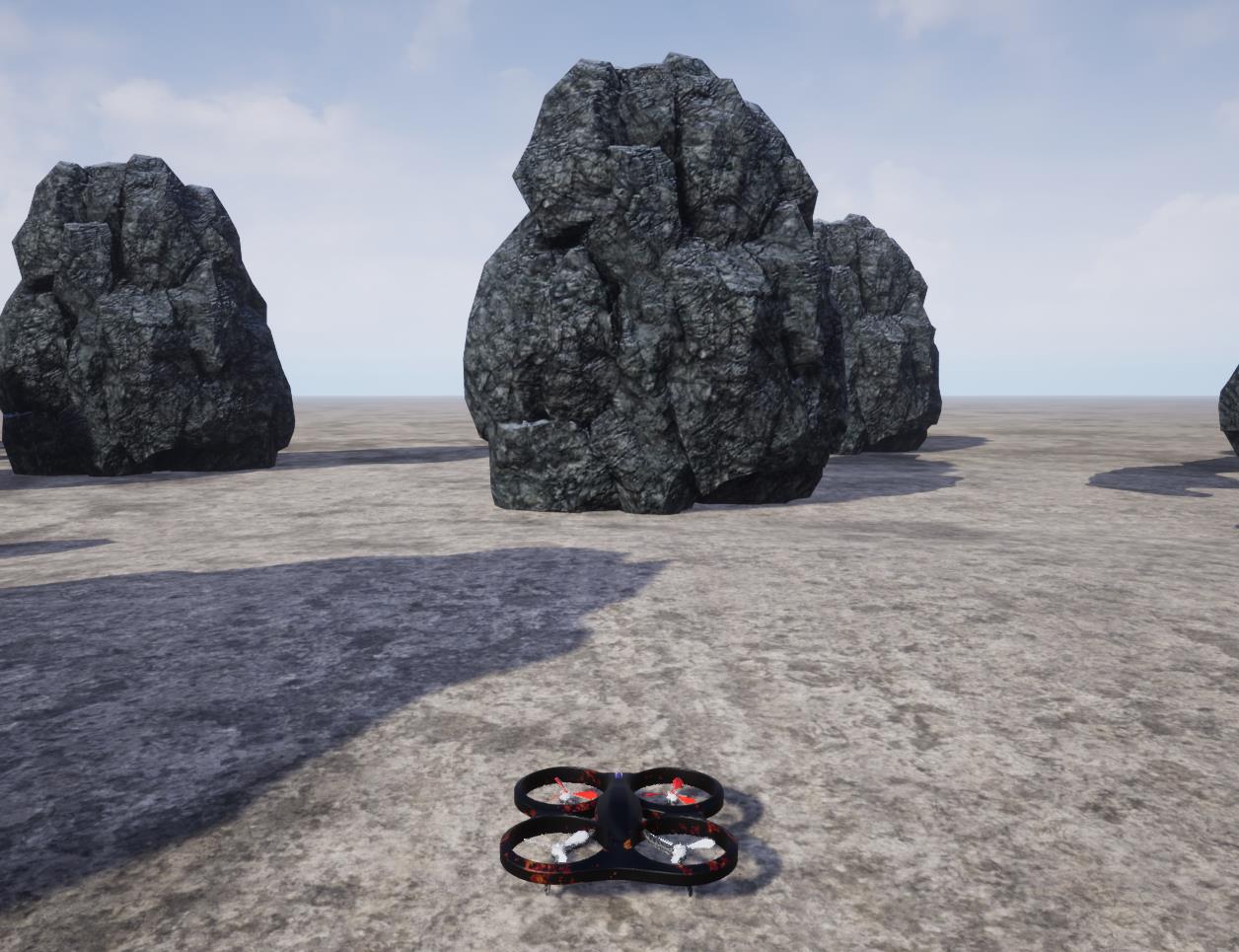}}
	\hfill
	\subfloat[Top view of the environment]{\includegraphics[width=0.24\textwidth]{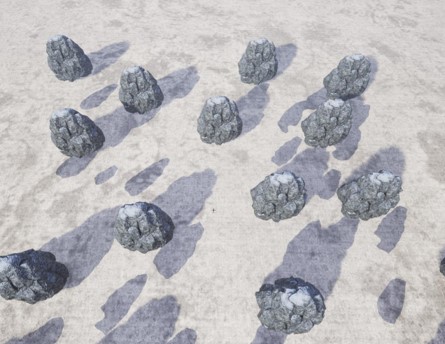}}
	\caption{Customized training environment created using Unreal Engine. The quadrotor takes off from the center of the environment. The goal is set randomly on the circle with a radius of 70 meters and centred on the take-off point. Some stones were randomly placed as obstacles.}
	\label{fig_training_env}
\end{figure}

The navigation network is trained from scratch in AirSim \cite{airsim2017fsr} simulator which is built on Unreal Engine. This simulator can provide high fidelity depth image and a low-level controller to stabilize the UAV. 
As shown in Fig. \ref{fig_training_env}, a customized environment is created for the training. The size of the environment is square with 200 meters on each side. Some stones were randomly placed as obstacles. At the beginning of each episode, the quadrotor takes off from the centre of the environment. The goal position is set randomly on the circle with a radius of 70 meters and centred on the take-off point. The episode terminated when the quadrotor reaches the goal position with an accept radius of 2 meters or crashed on the obstacles. At each time step, the neural network controller received the depth image as well as the velocity and position information of the quadrotor to generate the velocity setpoint in 3D environment as the control command. The controller is running at 10Hz and the velocity control is realised by the low-level controller provided by AirSim.

To get a smooth velocity command, we use continuous action space. An off-policy model-free reinforcement learning algorithm, Twin Delayed DDPG (TD3) \cite{fujimoto2018addressing}, is used for model training. 
As the successor of the DDPG method,  TD3 addresses the overestimate problem issue of Q-value in DDPG by introducing three critical tricks: clipped double Q-Learning, delayed policy update and target policy smoothing \cite{SpinningUp2018}. 
This DRL algorithm is widely used for continuous control problem. The hyper-parameters are tuned based on massive training. 
The final hyper-parameters of the algorithm are summarized in Table \ref{table_hyperparameters} in Appendix.

\subsection{Reward Function Design}
Reward function is critical for DRL problem. In general, the reward function for navigation is simple, we can only reward for reaching the goal position as soon as possible and punish for collision. However, because of the huge state space for the navigation task, especially in 3D environment, it's better to introduce some continuous reward signal to guide the exploration and speed up the training process. 
After a lot of testing, a hand-designed reward function is utilized, which consists of a continuous goal approaching reward and some penalty terms:

\begin{equation}
        r(s_t)=
        \begin{cases}
                10, &\text{if success} \\
                R_{goal} - P_{state}, &\text{otherwise}
        \end{cases}
\end{equation}
where $R_{goal} = d(s_{t-1}) - d(s_t)$ is the goal approaching reward and $d(s_t)$ is the \textit{Euclidean distance} from current position to goal position at time $t$. $P_{state}$ is the penalty term at current step:
\begin{equation}
    P_{state} = \omega_1 \cdot C_{obs} - \omega_2 \cdot  C_{act} - \omega_3 \cdot C_{pos}
\end{equation}
where $C_{obs}$, $C_{act}$ and $C_{pos}$ are penalty terms for obstacle, action, and position error.
\begin{equation} \label{eq_c_obs}
        C_{obs} = \frac{d_{safe} - d_{obs}(s_t)}{d_{safe} - d_{min}}
\end{equation}
is the penalty term to prevent the quadrotor from getting close to the obstacle. In equation \ref{eq_c_obs}, $d_{safe}$ and $d_{min}$ is the safety distance and minimum distance allowed to the obstacles. $d_{obs}(s_t)$ is the minimum distance to the obstacle at time $t$. In our training process, $d_{safe}=5$ and $d_{min}=1$, which means we give punishment if the quadrotor gets close to the obstacle in 5 meters. When the minimum distance to the obstacle is less than 1 meter, it is considered crashed and this episode terminates. To stabilize the training process, the continuous reward part is constrained to -1 to 1. 

\subsection{Training Result}
\begin{figure}
    \centering
    \subfloat[Episode reward]{\includegraphics[width=0.24\textwidth]{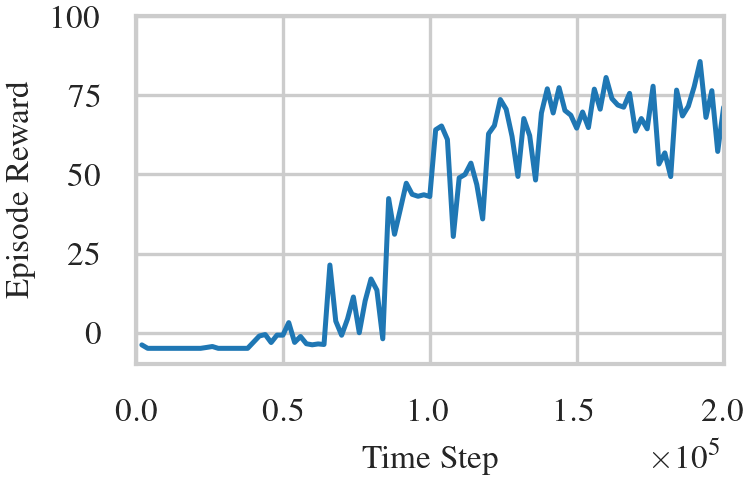}}
    \hfill
    \subfloat[Success rate]{\includegraphics[width=0.24\textwidth]{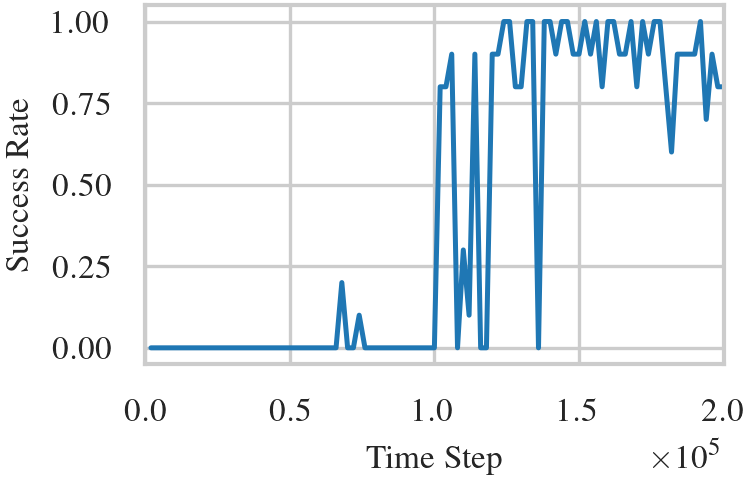}}
    \caption{Mean episode reward and success rate versus the training step curves. The success rate is obtained by evaluating each learned policy over 10 randomly generated navigation tasks without action noise. The evaluation is executed every 2k time steps during training.}
    \label{fig:training_result}
\end{figure}

The policy network is trained for 200k time steps (around 1000 episodes) in the simulation environment. To speed up the training process, the AirSim simulation clock speed is set to 10 makes it can run 10 times faster than real time. The total training process took about 7 hours on an PC with Intel i7-8700 processor and Nvidia GeForce GTX1060 GPU. The episode reward and success rate are plotted in Fig. \ref{fig:training_result}. From the training result, the policy gets about 75\% success rate when the algorithm converged, which means that the network can guide the UAV to the goal position without collision with any obstacles in most of the scenario. 

\section{Post-hoc Explanation Method}
In this section, we introduce our model explanation method. To keep the network performance, post-hoc explanation approach is used to explain the trained network. Feature attribution is a common method to get post-hoc model explanation. In most cases, gradients can be used to reflect the feature attribution if the model is differentiable. However, because of the model saturation and discontinuities, gradients is not always fear to all the features. In our work, a novel feature attribution metrics, SHAP value, is used to measure the feature attribution. SHAP value is provably the only distribution with certain desirable properties, which can make the explanation fairer. 

Different from the traditional image classification or purely vision based navigation, in our case, the input of the network consists both depth information (image) and state information (scalar). Hence, the navigation network consists of a CNN perception part to deal with the image information and a fully connected part to fuse the image feature and state feature. Because of this specific kind of network architecture, our explanation consists of both visual part for the image input and texture part for state features. 

\subsection{Visual explanation}
In our problem, depth image provide the obstacle information. A CNN part is used to extract the visual feature from the raw depth image. So CNN visualization is important for understanding the output of the learned policy. 

Understanding the insights of CNN has always been a pain point, though CNN can get excellent predictive performance. In \cite{zeiler2014visualizing}, a deconvolutional network (Deconvnet) approach was proposed to visualize activated pattern in each hidden unit. This method can visualize features individually but is limited as it is hard to summarize all hidden patterns into one pattern. Simonyan \emph{et~al} \cite{simonyan2013deep} visualize partial derivatives of predicted class scores w.r.t.pixel intensities, while Guided Backpropagation \cite{springenberg2014striving} makes modifications to ‘raw’ gradients that result in qualitative improvements. This method can provide fine-grained visualizations. 

In \cite{Zhou_2016_CVPR}, the authors proposed Class Activation Map (CAM) using global average pooling (GAP) layer to summarize the activation of the last CNN layer. However, it is only applicable to a particular CNN architecture where the GAP layer is fed directly into the soft-max layer. To address this problem, Grad-CAM \cite{selvaraju2017grad} method combined feature maps and the gradient signal that does not require any modification in the network architecture . It can be used to off-the-shelf CNN architecture. Grad-CAM uses the gradient information flowing into the last convolutional layer of CNN to assign importance values to each neuron for a particular decision of interest.
 
To visualize the CNN perception part of our network, a method combined both CAM and SHAP values is proposed. Similar to CAM method, global average pooling (GAP) layer is reserved to summarize the visual feature in our CNN perception network. The output of the GAP layer is defined as the CNN feature. 
Different from CAM and Grad-CAM, in our method, the SHAP value of CNN feature is used to determine the importance of the CNN feature which generated from the corresponding activation map. 
A coarse localization map highlighting the important regions in the image is generated by a weighted sum of the last CNN activation map, where SHAP value is the weight. 

\begin{figure}
	\centering
	\includegraphics[width=0.45\textwidth]{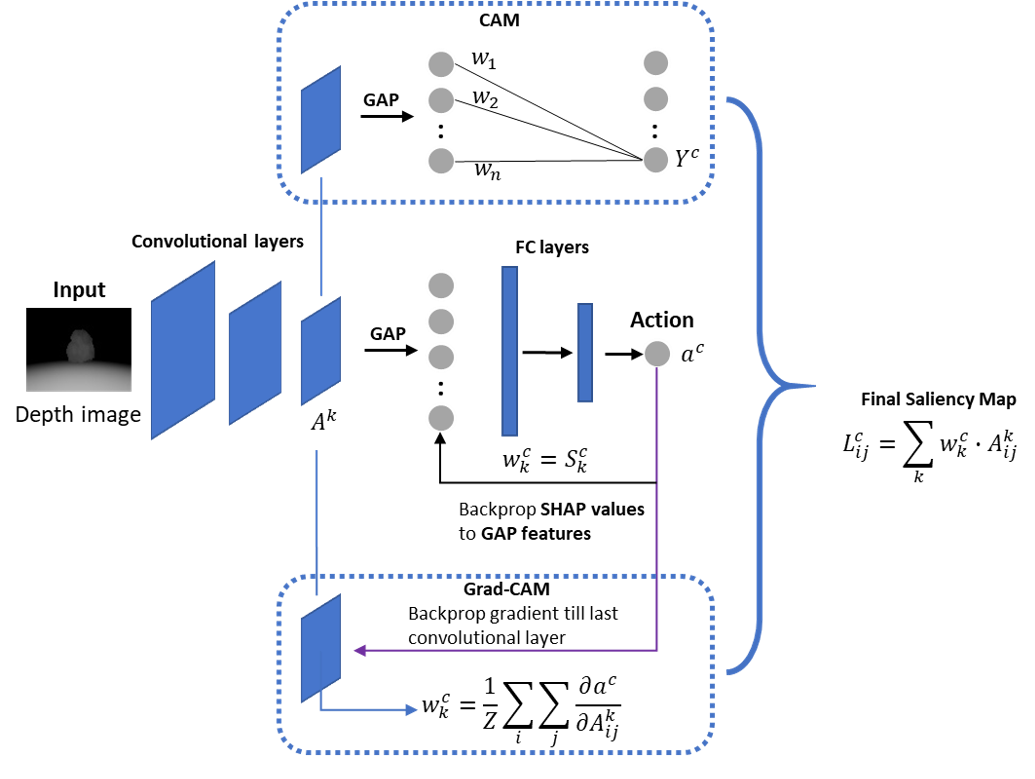}
	\caption{SHAP-CAM method. Different from CAM and Grad-CAM, in our problem, the network output is action rather than class score. We use Global Average Pooling as CAM to get the CNN perception feature intensity. Then SHAP value is calculated directly as the weight of the saliency map.}
	\label{fig:SHAP-CAM}
\end{figure}

Comparing to Grad-CAM, SHAP value is used as weights of the forward activation maps rather than gradients because SHAP value has some unique properties comparing to the gradient, such as efficiency which means the feature attributions should sum to the prediction value. So using SHAP values as the feature importance can provide a fairer attribution of the activation maps. The difference between CAM, Grad-CAM and our method is shown in Fig. \ref{fig:SHAP-CAM}.

\subsection{Texture explanation}
In addition to the visual part, our model also takes some state features as input of the network. To get a reasonable explanation of the model output, both image and state input should be considered. To explain the state feature contribution, some texture explanations are provided based on their SHAP values. 

\begin{figure}
    \centering
    \includegraphics[width=0.4\textwidth]{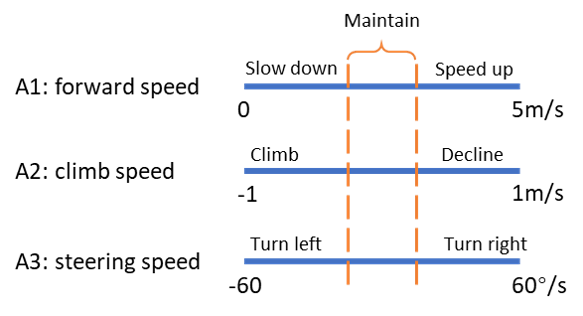}
    \caption{Action description. Each action is divided into 3 parts. While the prediction fall into the central part, we say it is maintain the current state. Otherwise, there will be a textual description of each action. The final description will be the combination of these three individual descriptions.}
    \label{fig:action_description}
\end{figure}

Our model has 3 continuous action outputs, horizontal speed $v_{xy}^{cmd}$, vertical speed $v_z^{cmd}$ and steering angular speed $\phi^{cmd}$. To get the textual explanation, each action is divided into 3 parts based on the reference action, as shown in Fig. \ref{fig:action_description}. If the action is similar to the reference action, we say that this action is to maintain current action. If the output action either bigger or smaller than the reference action, a specific text is used to describe the action, such as 'slow down' or 'speed up' for the horizontal speed $v_{xy}^{cmd}$. The final textual output of the action is the combination of these three action textual descriptions, for example, the action can be described as 'slow down, maintain the altitude and turn right'. 

Finally, with both visual and textual explanation, every output of the network can be explained to non-expert users to illustrate the reason of the network's decision. These explanations only take one forward propagation. Hence, it can also provide real-time explanation during flight. 

\section{Model Explanation}

In this section, we explain the trained model using the proposed explanation method. Both visual and texture explanation are provided to analysis every model prediction which can provide real-time action explanations for non-expert users. The visual part shows the attention of the CNN perception part, and the texture part summarises the contribution of other state features. In addition, activation map of the last CNN layer is drawn to show the visual feature extracted by the CNN part. Finally, to help expert to diagnose and improve the network design, some global explanations are also provided to analyse the network performance based on the data gathered in 20 continuous episodes.

\subsection{Defining the Reference Input}
Baselines or norms are essential to all explanations \cite{kahneman1986norm}. 
Feature attribution method always generates the contribution of each feature based on a reference input or baseline input. Thus, the choice of the reference input is critical for obtaining insightful results \cite{shrikumar2017learning}. In practice, choosing a good reference would rely on domain-specific knowledge. For instance, in object recognition networks, it is the black image. 

\begin{figure}
    \centering
    \includegraphics[width=0.2\textwidth]{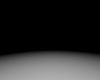}
    \caption{Reference depth image. In our case, we choose the depth image at the target flight height without any obstacles as the reference image input.}
    \label{fig_ref_depth_img}
\end{figure}

In our case, we choose the depth image at the target flight height without any obstacles as the reference image input. For state feature input, we set the $o_\text{ref} = [d_{xy}=70, d_z=0, \xi=0, v_{xy}=0, v_z=0, \phi=0$] which means the UAV just take off from the start point and has no velocity. The reference image is shown in Fig. \ref{fig_ref_depth_img}. Based on this reference input, we can get reference model output from the trained network: $v_{xy}^\text{cmd} = 3.71 m/s, v_z^\text{cmd}=-0.03 m/s, \phi^\text{cmd} = 4.15 ^{\circ}/s$, which means the network wants the quadrotor to speed up and turn a little bit right based on the reference input.

\subsection{Local explanation}
\begin{figure*}
	\centering
	\subfloat[Action explanation at $t=0$]{\includegraphics[width=0.3\textwidth]{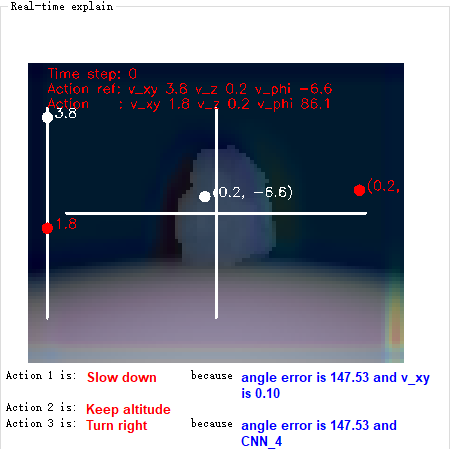}}
	\hfill
	\subfloat[Action explanation at $t=53$]{\includegraphics[width=0.3\textwidth]{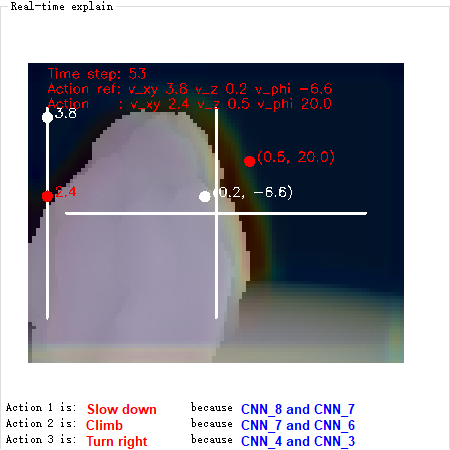}}
	\hfill
	\subfloat[Action explanation at $t=89$]{\includegraphics[width=0.3\textwidth]{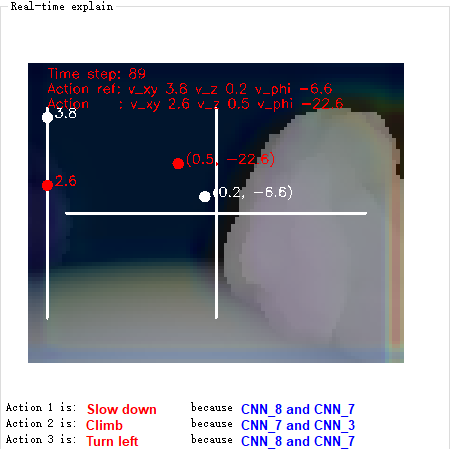}}
	\caption{Action explanation at 3 different time steps. At $t=0$, the action is slow down, keep altitude and turn right. Mainly because the big angle error. At $t=53$, the action is slow down, climb and turn right, mainly because the image features. From the heat map, we can see the quadrotor is close to the stone and the CNN detected the edge of the stone. At $t=89$, the action is slow down, climb and turn left. This is also cause of the image feature.}
	\label{fig:action_explain}
\end{figure*}

Local explanation can be generated for every time step and every model prediction. Here, 3 specific time steps are choosing to demonstrate our visual and textual explanation for actions in one of the model evaluation episode. 
As shown in Fig. \ref{fig:action_explain}, at $t=0$, the action is slow down, keep altitude and turn right. 
The explanation shows both slow down and turn right are caused by the angular error to goal. 
This makes sense because the direction at $t=0$ doesn't match the goal position, so the UAV need turn right. 
At $t=53$, the action is slow down, climb and turn right. The explanation shows this is caused by the CNN feature. From the heat-map generated using SHAP-CAM, we can see the CNN detected left edge of the stone which is the obstacle. 
At $t=89$, the action is slow down, climb and turn left. This is also caused by the CNN feature.

\begin{figure*}
    \centering
    \subfloat[Last CNN layer activation map at $t=53$]{
        \includegraphics[width=\textwidth]{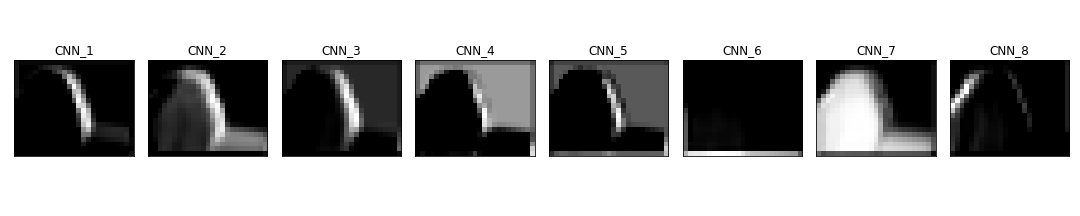}
    }
    \hfill
    \subfloat[Last CNN layer activation map at $t=89$]{
        \includegraphics[width=\textwidth]{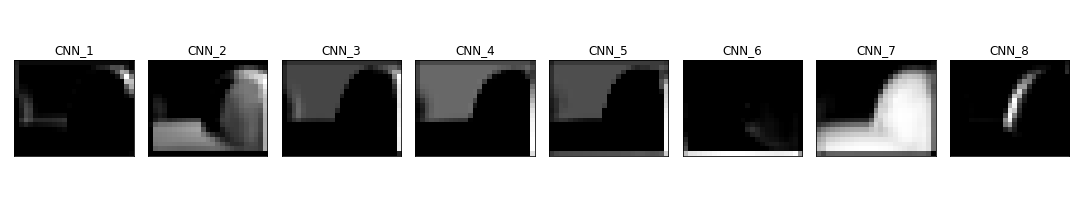}
    }
    \caption{Last CNN layer activation map. From this map we can get the meaning of different CNN feature. For example, according to Fig. \ref{fig:action_explain}, at $t=53$, the action 3 is turn right, because CNN\_4 and CNN\_3 feature. From Fig. \ref{fig:last_cnn_map} (a), CNN\_3 and CNN\_4 is the right edge of the stone.}
    \label{fig:last_cnn_map}
\end{figure*}

To find out the meaning of the CNN features, we also plotted the last CNN layer activation map at both $t=53$ and $t=89$ as shown in Fig. \ref{fig:last_cnn_map}. From this activation map, we can see at $t=53$, CNN feature 8 is the left and right edges of the obstacle which contributes most to the slow down action. CNN feature 7 is the obstacle and some ground which contributes to the climb. CNN feature 4 shows the right side edge of the obstacle with some free space background, which leads to the turn right action.

\begin{figure*}
    \centering
    \includegraphics[width=\textwidth]{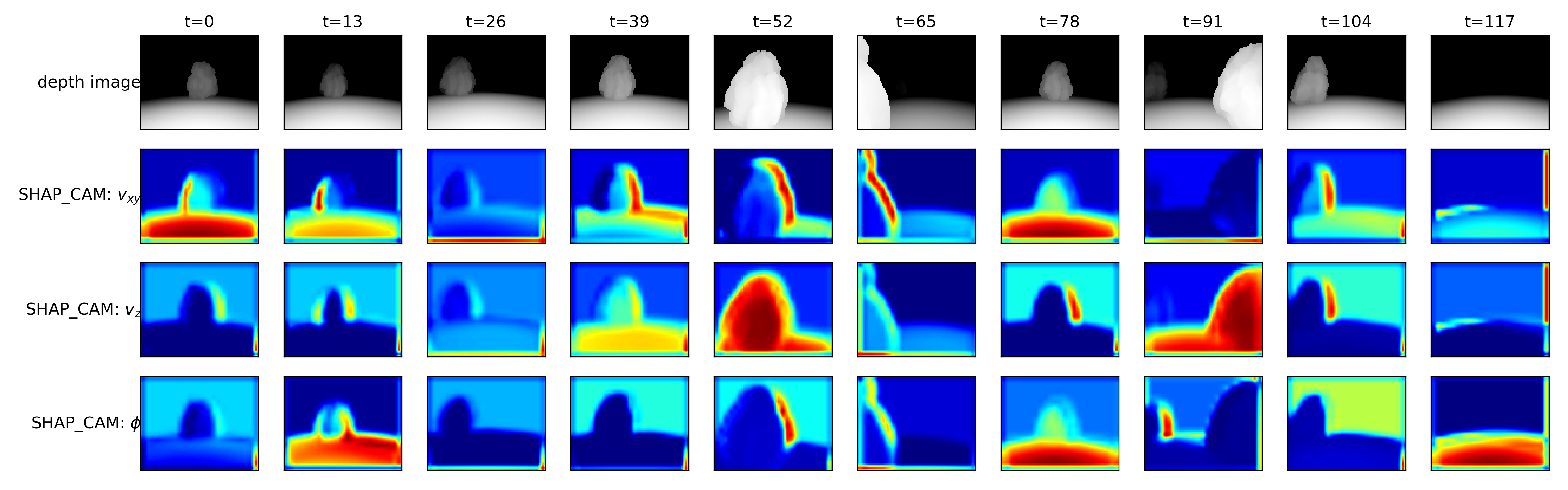}
    \caption{Depth image and SHAP-CAM at 10 different time steps in the evaluation episode. The first line is the input depth image. The second to fourth lines are three SHAP-CAM activation maps for three network outputs.}
    \label{fig_shap_cam_traj}
\end{figure*}

\begin{figure*}[t]
	\centering
	\includegraphics[width=\textwidth]{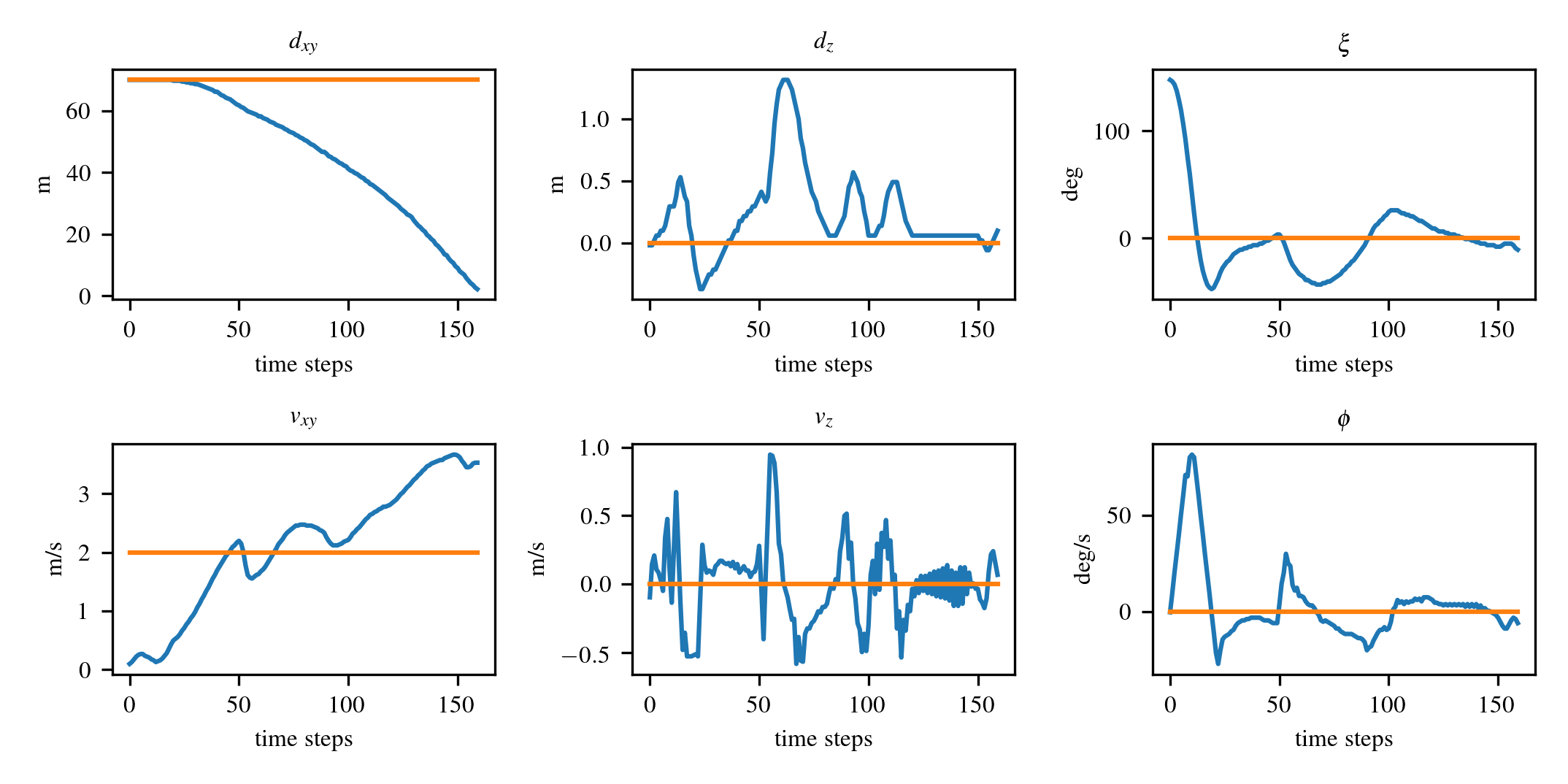}
	\caption{State feature in the evaluation episode. Blue line is the state feature and orange line is the reference state feature value.}
	\label{fig_state_traj}
\end{figure*}

\begin{figure*}[t]
    \centering
    \includegraphics[width=\textwidth]{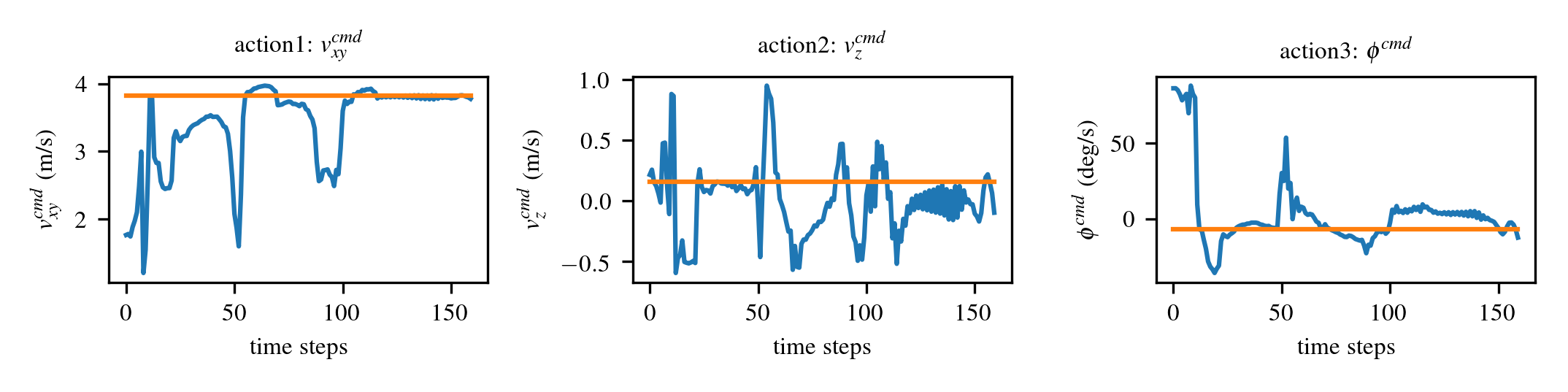}
    \caption{Network output in the evaluation episode. Blue line is the action and orange line is the reference action.}
    \label{fig_action_traj}
\end{figure*}

To illustrate more local explanation result, we choose one episode from the evaluation process and explain the model predictions at each time step using SHAP-CAM. Fig. \ref{fig_shap_cam_traj} shows the depth image and the SHAP-CAM activation map for 3 actions at different time steps. From Fig. \ref{fig_shap_cam_traj}, we can see that at different time step, the network decision-making process for different output is rely on the different visual pattens. Moreover, Fig. \ref{fig_state_traj} and Fig. \ref{fig_action_traj} show the control command and state features during this evaluation episode. From $d_{xy}$ in Fig. \ref{fig_state_traj}, we can see that the UAV always fly towards the goal position and the distance to goal is reducing over the trajectory. Finally, at $t=160$, UAV reached the goal position. 

\subsection{Global explanation}
\begin{figure*}[t]
    \centering
    \includegraphics[width=0.95\textwidth]{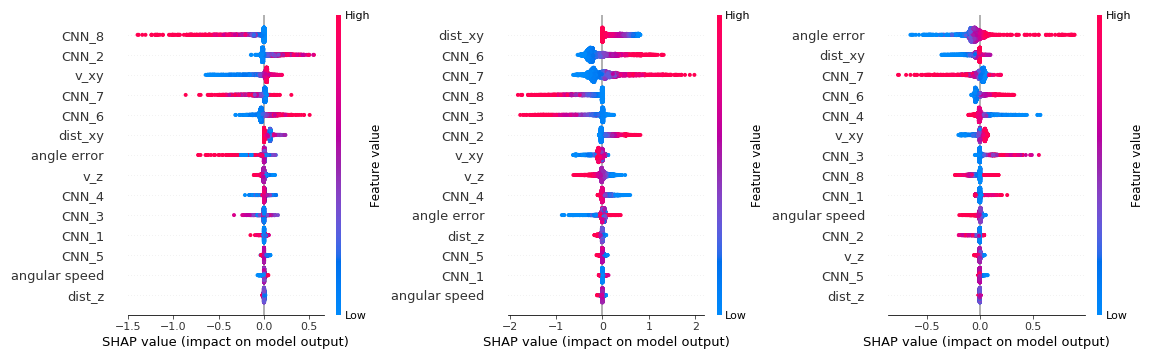}
    \caption{Feature analysis over 20 trajectories for 3 actions, forward speed $v_{xy}^\text{cmd}$ (left), vertical speed $v_z^\text{cmd}$ (middle) and steering speed $\phi^\text{cmd}$ (right). For each action, all the features are sorted according to their average attribution. For $v_{xy}^\text{cmd}$, the CNN\_8 and CNN\_2 feature contribute most, then the current forward speed v\_xy. For $v_z^\text{cmd}$, the distance to goal contributes most. For $\phi^\text{cmd}$, the angle error to goal is the most important feature.}
    \label{fig_feature_analysis}
\end{figure*}

Except for the local explanation, some global explanations are provided. We summarized all the feature attribution over the 20 trajectories, 2858 time steps in total. Fig. \ref{fig_feature_analysis} shows the SHAP summary plot that orders the features based on their importance to the different action. We can see that the CNN feature contributes most to action $a1: v_{xy}^\text{cmd}$ and $a2: v_z^\text{cmd}$. Except the CNN features, the current horizontal velocity $v_{xy}$ and distance to goal $d_{xy}$ are the most importance features contribute to $a1: v_{xy}^\text{cmd}$. $d_{xy}$, $v_{xy}$ and $v_z$ contributes more to $a2: v_z^\text{cmd}$, the vertical velocity command. The angle error $\xi$ is the most important feature to $a_3: \phi^\text{cmd}$. 

\begin{figure}[t]
    \centering
    \includegraphics[width=0.45\textwidth]{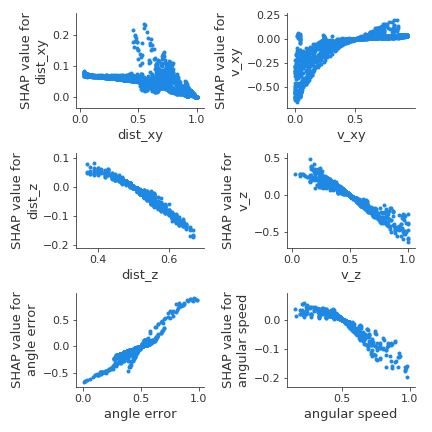}
    \caption{Feature dependence plot using 2858 sample from 20 trajectories. The x-axis is the feature value, the y-axis is its SHAP value. The feature value is normalized to 0 to 1 so angle error is 0.5 means $\xi=0$. The first row shows the SHAP value of state feature $d_{xy}$ and $v_{xy}$ with respect to $a1: v_{xy}^\text{cmd}$. The second row shows the SHAP value of state feature $d_z$ and $v_z$ with respect to $a2: v_z^\text{cmd}$. The third row shows the SHAP value of state feature $\xi$ and $\phi$ with respect to $a_3: \phi^\text{cmd}$.}
    \label{fig_feature_dependence}
\end{figure}

With the feature value and its SHAP value, we can invest the relationship between the feature intensity and its importance measurement as shown in Fig. \ref{fig_feature_dependence}. From the plot, we can find that there is some relationship between the feature value and the SHAP value. For example, from the first plot in the third row of Fig. \ref{fig_feature_dependence}, which shows the SHAP value of state feature $\xi$ with respect to $a_3: \phi^\text{cmd}$. From these two plots we can see that the angle error shows a positive correlation to its SHAP value. However, the angular speed shows a negative correlation.

\section{Real World flight test}
To validate the performance of our reactive navigation controller, some real world outdoor experiments are carried out.

\subsection{Flight Platform}
A self-assembled quadrotor platform is used to evaluate the trained navigation network, as shown in Fig. \ref{fig:flight_path_real} (a). The platform is based on the S500 quadrotor framework, equipped with a Pixhawk flight controller for the low-level attitude and velocity control, which is also providing the position and velocity information of the quadrotor. A Intel RealSense D435i camera is mounted forward to perceive the depth information in front of the quadrotor. The on-board computer is a Nvidia Jetson Nano. It is used to run the neural network and generate the velocity command signals. The velocity command signal is sent to the flight controller as velocity setpoint via serial port at 10 Hz. 


\subsection{Model Retraining}
\begin{figure}
	\centering
	\subfloat[Episode reward]{\includegraphics[width=0.24\textwidth]{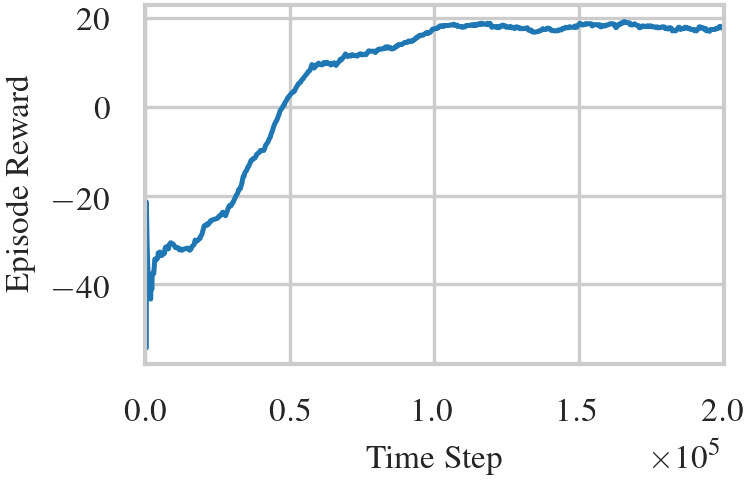}}
	\hfill
	\subfloat[Success rate]{\includegraphics[width=0.24\textwidth]{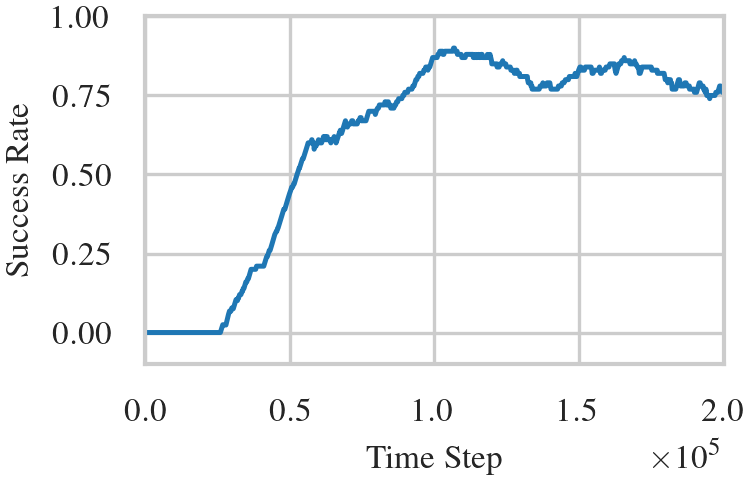}}
	\caption{Training result in gazebo simulation environment. The model is trained from scratch for 20k time steps. The episode reward and success rate is smoothed over 100 episodes. After training, the model can achieve about 75$\%$ success rate.}
	\label{fig:training_result_gazebo}
\end{figure}

To simplify the experiment, we fixed the the height of the quadrotor during flight to 5m and reduce the controller output to only forward velocity and steering velocity. Notable, to reduce the gap between the simulation and real environment, the network was retrained in a custom Gazebo environment. The new environment uses simulated trees as obstacles. The controller is running in the PX4 Simulation in the Loop (SITL) configuration. Also, to keep safety of the quadrotor, the maximum forward velocity is limited to 1m/s. The network is trained for 20k timesteps. Training result is shown in Fig. \ref{fig:training_result_gazebo}. The success rate is about 75$\%$ after training.

\subsection{Real World Fight}
\begin{figure}
	\centering
	\subfloat[Real world test environment\label{fig_path_a_env}]{\includegraphics[width=0.4\textwidth]{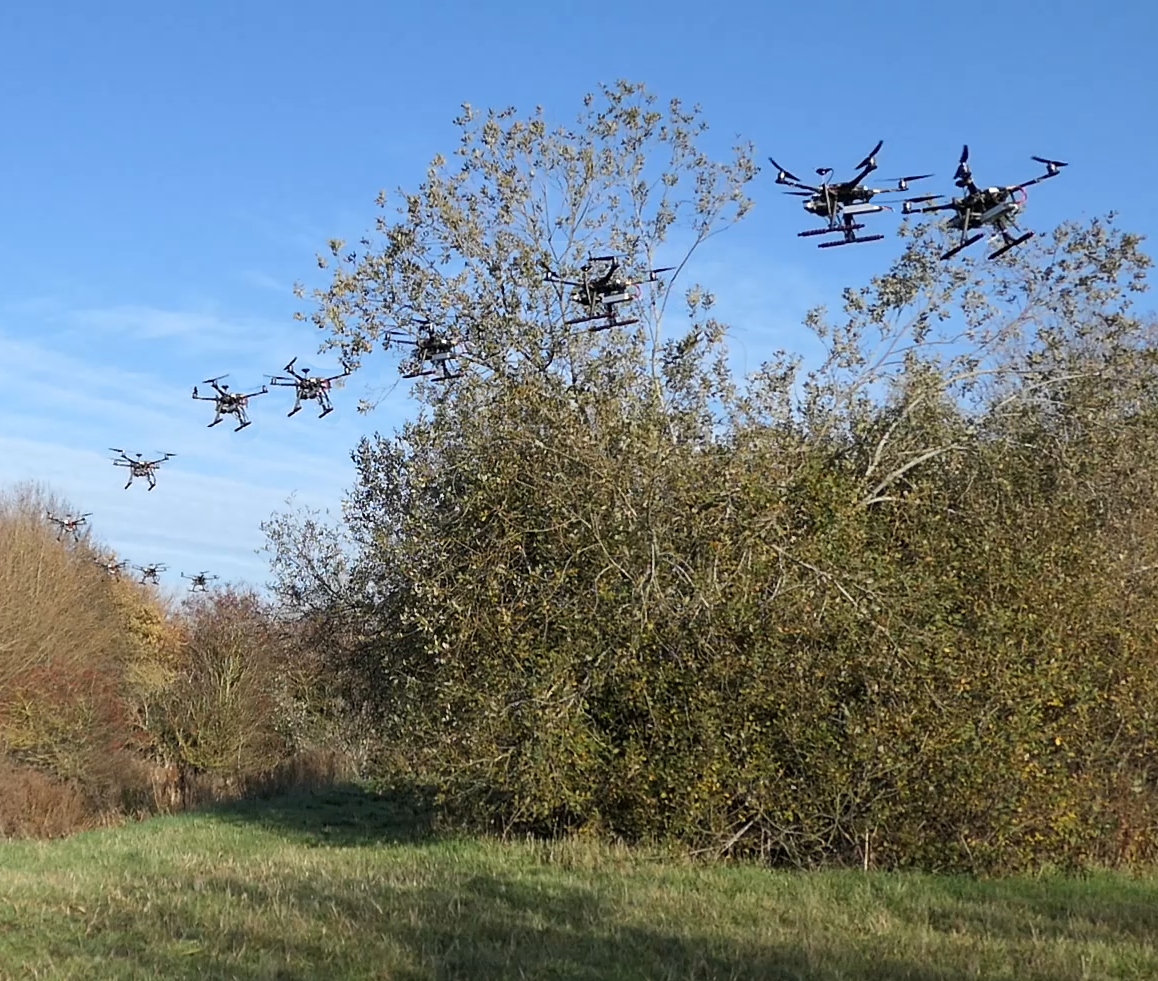}}
	\hfill
	\subfloat[Test flight path\label{fig_path_b_path}]{\includegraphics[width=0.49\textwidth]{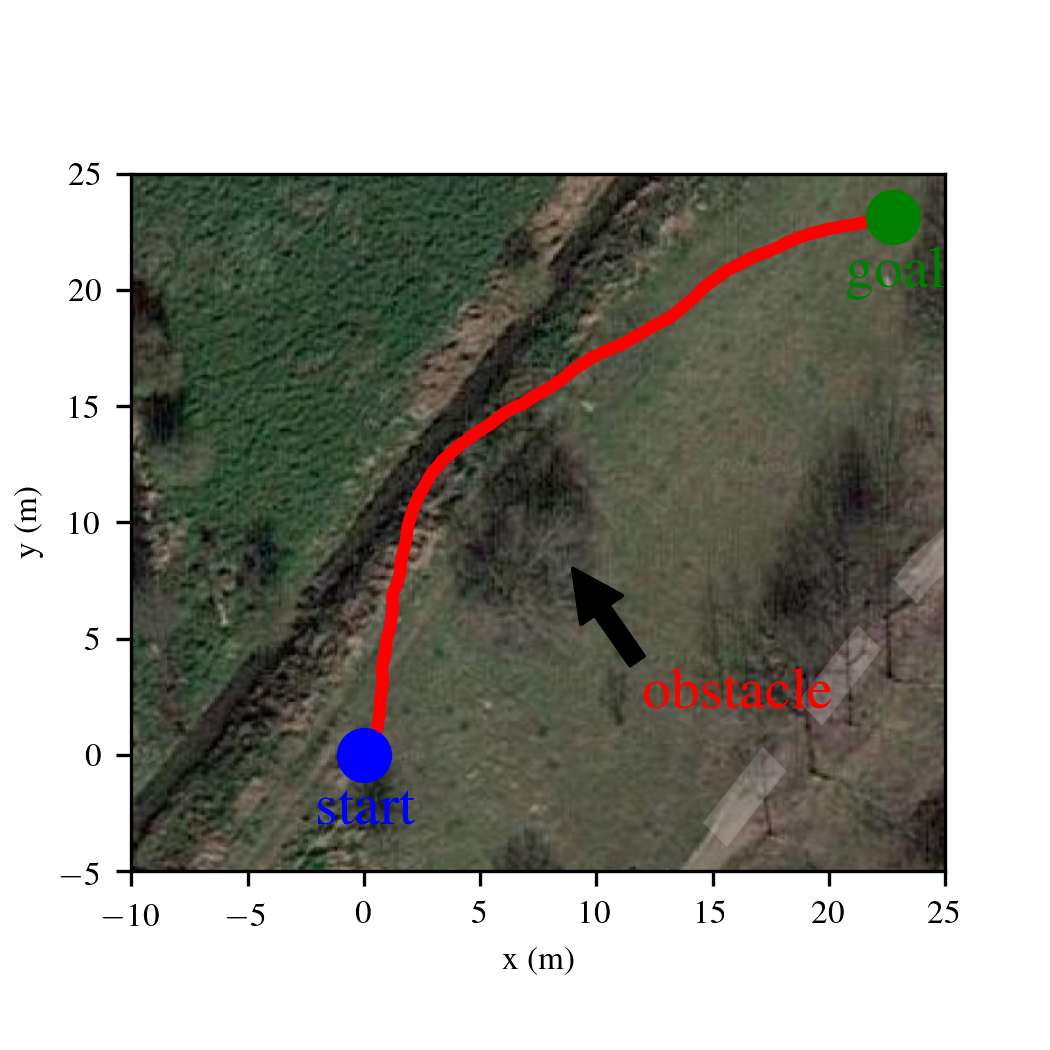}}
	\caption{Real world flight test. Fig.  shows the test environment. The quadrotor is facing a tree as obstacles in front of it. The goal position is set about 35m behind the obstacle. and the flight trajectory.}
	\label{fig:flight_path_real}
\end{figure}

After training, the trained network is deployed directly to the real flight platform. The real-world test environment is shown in figure \ref{fig_path_a_env}. A big tree was chose as obstacle and the goal position was set behind the tree. One of the flight paths is shown in figure \ref{fig_path_b_path}. From the flight path, the trained reactive controller can navigate the quadrotor to avoid the obstacle and reach the goal finally. 

As for comparison, we also tested a traditional search based obstacle avoidance algorithm. PX4 avoidance system which is based on the 3DVFH+ is deployed on the test platform. This local planner plans in a vector field histogram include some history information. The flight results show both algorithm can navigate the quadrotor to the final position. However, using the same hardware, the PX4 avoidance system can only run at 10Hz. In contrast, our neural network-based reactive controller can run at 60Hz maximum. This shows the computational advantage of the reactive controller rather than conventional planner, which is benefit for lightweight UAV with limited computation resources. 

\begin{figure}
    \centering
    \includegraphics[width=0.48\textwidth]{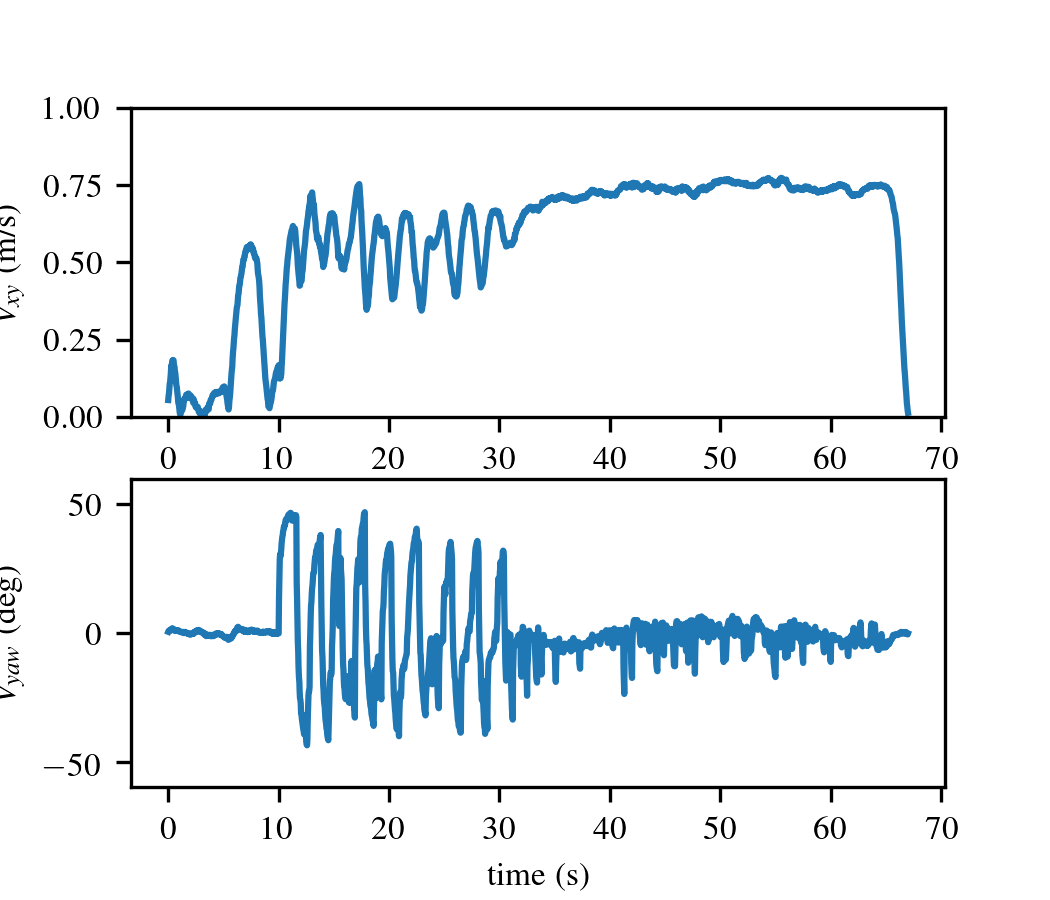}
    \caption{Forward speed (up) and steering speed (down) during the real flight test.}
    \label{fig:action_real}
\end{figure}

However, according to the flight test result, some problems have been exposed. The main problem is the output oscillation. As shown in Fig. \ref{fig:action_real}, the output during the avoiding process is not very smooth in the real test, although this output is smooth in the simulation. We think this is caused by the persistent gap between the simulation and real environment, such as the difference dynamics and the different input image. This also reflect the domain adjustment problem for all the learning-based control system. 

\section{Conclusion}
In this paper, the UAV autonomous navigation problem is solved with the DRL technique. Different from other works, this paper focused on improving the model explainability rather than treat the trained model as a black box. Based on the feature attribute, both visual and textual explanation are generated to open the black box. 
To get a better visual explanation of the CNN perception part, a new saliency map generation method proposed combining both CAM and SHAP values. 
Our method can provide both visual and textual explanation for non-expert users of every model output, which is important for the application of DRL based model in the real world. 

Because this paper mainly focused on the explanation part, the trained model is not perfect enough. There still some explanations don't make sense. In the future, the model will be fine-trained and improved based on the explanation.  

Feature attribution method can provide some explanation of the deep neural network, however, it is still pretty shallow. For example, attributions do not explain how the network combines the features to produce the answer and why gradient descent converged. In the future, we will looking for other method to make better explanation to the trained network. Also, we are still interested in the explanation of the training process of DRL to reflect the knowledge acquisition process.

\appendix
\subsection{Hyperparameters of TD3}
The hyperparameters are shown in Table 

\begin{center}
    \begin{table}[t]
        \caption{Hyperparameters of TD3}
        \label{table_hyperparameters}
        \begin{center}
                \begin{tabular}{cc}
                        \hline
                        \textbf{Hyperparameter} & \textbf{Value} \\
                        \hline
                        mini-batch size & 128 \\ 
                        replay buffer size & 50000 \\
                        discount factor & 0.99 \\
                        learning rate & 0.0003 \\
                        random exploration steps & 2000\\
                        square deviation of exploration noise & 0.3 \\
                        \hline
                \end{tabular}
        \end{center}
\end{table}
\end{center}

\ifCLASSOPTIONcaptionsoff
  \newpage
\fi

\bibliographystyle{IEEEtran}
\bibliography{./ref}


\end{document}